\pdfoutput=1

\documentclass[11pt]{article}

\usepackage{emnlp2021}

\usepackage{times}
\usepackage{latexsym}

\usepackage[T1]{fontenc}

\usepackage[utf8]{inputenc}

\usepackage{microtype}

\usepackage{times}  
\usepackage{helvet} 
\usepackage{courier}  
\usepackage{graphicx} 
\usepackage{subcaption}
\usepackage{natbib}  
\usepackage{caption} 
\usepackage{amsmath}
\usepackage{booktabs}
\usepackage{xcolor}
\usepackage{xspace}
\usepackage{enumitem}
\usepackage{multirow, multicol}
\usepackage{flushend}

\usepackage{amsfonts}
\newcommand{\abr}[1]{\textsc{#1}}
\newcommand{\dn}{\abr{simmc }{\small 2.0}\xspace} 

\newcommand{\simmc}{SIMMC\xspace}

\newcommand{\reffig}[1]{Fig.~\ref{#1}}
\newcommand{\refsec}[1]{Sec.~\ref{#1}}
\newcommand{\reftab}[1]{Tab.~\ref{#1}}

\newcommand{\reportval}[2]{$#1$\footnotesize $\pm #2$}

\usepackage{booktabs,arydshln}
\makeatletter
\def\adl@drawiv#1#2#3{%
        \hskip.5\tabcolsep
        \xleaders#3{#2.5\@tempdimb #1{1}#2.5\@tempdimb}%
                #2\z@ plus1fil minus1fil\relax
        \hskip.5\tabcolsep}
\newcommand{\cdashlinelr}[1]{%
  \noalign{\vskip\aboverulesep
           \global\let\@dashdrawstore\adl@draw
           \global\let\adl@draw\adl@drawiv}
  \cdashline{#1}
  \noalign{\global\let\adl@draw\@dashdrawstore
           \vskip\belowrulesep}}
\makeatother

\belowdisplayskip \abovedisplayskip

\newlength{\sectionReduceTop}
\newlength{\sectionReduceBot}
\newlength{\subsectionReduceTop}
\newlength{\subsectionReduceBot}
\newlength{\abstractReduceTop}
\newlength{\abstractReduceBot}
\newlength{\captionReduceTop}
\newlength{\captionReduceBot}
\newlength{\subsubsectionReduceTop}
\newlength{\subsubsectionReduceBot}

\newlength{\eqnReduceTop}
\newlength{\eqnReduceBot}

\newlength{\horSkip}
\newlength{\verSkip}

\newlength{\figureHeight}
\usepackage{graphicx} 
\usepackage{multicol}
\usepackage{booktabs} 
\usepackage{algorithm} 
\usepackage{algorithmic} 
\usepackage{color}
\usepackage{amsmath}
\usepackage{amssymb}
\usepackage{amsfonts}
\usepackage{multirow, varwidth}
\usepackage{stfloats} 
\usepackage{verbatim}
\makeatletter
\newif\if@restonecol
\makeatother
\usepackage{xr}
\usepackage[amsmath,thmmarks]{ntheorem}

\usepackage{xspace}
\makeatletter
\DeclareRobustCommand\onedot{\futurelet\@let@token\@onedot}
\def\@onedot{\ifx\@let@token.\else.\null\fi\xspace}

\def\eg{\emph{e.g}\onedot} 
\def\ie{\emph{i.e}\onedot}

\def\etc{\emph{etc}\onedot}

\makeatother

\newcommand{\smtodo}[1]{}  
\newcommand{\sm}[1]{{\color{black}{#1}}}

\newcommand{\sktodo}[1]{}  
\newcommand{\sk}[1]{{\color{black}{#1}}}

\usepackage[misc]{ifsym} 
\newcommand*\samethanks[1][\value{footnote}]{\footnotemark[#1]}

\title{SIMMC 2.0: A Task-oriented Dialog Dataset\\for Immersive Multimodal Conversations}

\author{ \\
  \textbf{Satwik Kottur}\thanks{\hspace{5pt}Joint first authors},
  \textbf{Seungwhan Moon}\samethanks,
  \textbf{Alborz Geramifard},
  \textbf{Babak Damavandi} \\
  Facebook Reality Labs \& Facebook AI \\
  {\small \Letter}\hspace{3pt} \texttt{\{skottur,shanemoon,alborzg,babakd\}@fb.com} \\
}
\date{}

\begin{document}
\maketitle
\begin{abstract}
Next generation task-oriented dialog systems need to understand conversational contexts with 
their perceived surroundings, to effectively help users in the real-world multimodal 
environment. 
Existing task-oriented dialog datasets aimed towards virtual assistance fall short and do not 
situate the dialog in the user's multimodal context.
To overcome, we present a new dataset for Situated and Interactive Multimodal
Conversations, \dn, which includes $11K$ task-oriented user$\leftrightarrow$assistant 
dialogs ($117K$ utterances) in the shopping domain, grounded in immersive and photo-realistic
scenes.

The dialogs are collected using a two-phase pipeline:
(1) A novel multimodal dialog simulator generates simulated dialog flows, with an emphasis on 
diversity and richness of interactions,
(2) Manual paraphrasing of the generated utterances to collect diverse referring expressions.
We provide an in-depth analysis of the collected dataset, and describe in detail the four main benchmark tasks we propose.
Our baseline model, powered by the state-of-the-art language model, shows promising 
results, and highlights new challenges and directions for the community to study
\footnote{Code \& data are made publicly available: \url{https://github.com/facebookresearch/simmc2}}.
\end{abstract}

\section{Introduction}
\begin{figure}[t]
  \centering \includegraphics[width=0.99\columnwidth]{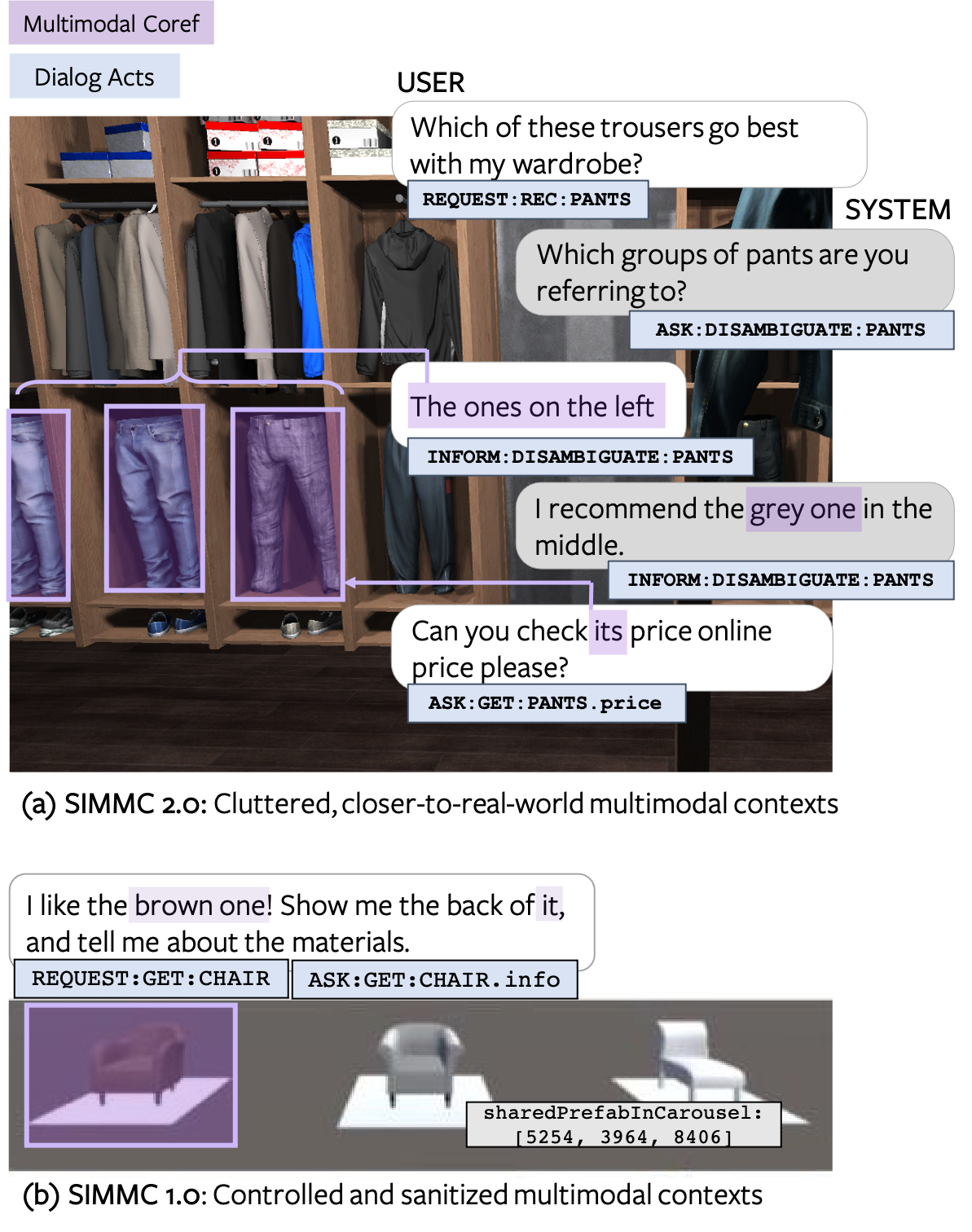}
  \vspace*{\captionReduceTop}
  \caption{Illustration of a Situated Interactive Multimodal Conversation (SIMMC), which presents a task-oriented user$\leftrightarrow$assistant dialog grounded in a co-observed multimodal context. The newly collected \dn{} dataset includes complex and photorealistic multimodal contexts, which poses more challenges for the Multimodal Coreference Resolution task (MM-Coref) and the Multimodal Dialog State Tracking task.}
  \vspace*{\captionReduceBot}
  \label{fig:teaser}
\end{figure}

The Situated and Interactive Multimodal Conversational AI (SIMMC) challenge \cite{moon2020situated}, held as part of DSTC9 \cite{dstc9}, aimed to lay the foundations for the real-world assistant agents that can handle multimodal inputs, and perform multimodal actions.
Specifically, it provided the SIMMC datasets as new benchmarks for studying task-oriented
dialogs that encompass a situated multimodal user context in the form of a co-observed image 
or virtual reality (VR) environment.
Since the introduction of the dataset, a number of follow-up work \cite{simmc1-team1,simmc1-team2,simmc1-team3,simmc1-team4,simmc1-team5} have established a new set of state-of-the-art baselines for the multimodal task-oriented dialog systems on SIMMC.

\sk{
Though SIMMC serves as a step towards building multimodal virtual agents, the dataset
falls short (understandably so) in the complexity of the considered multimodal contexts.
In particular, the co-observed image or VR environment is simplistic and far from realistic
user situations.
To bridge this gap, we take inspiration from the first SIMMC challenge \cite{moon2020situated}
and}
\sm{
propose a new multimodal dialog dataset (\dn) for the community to tackle and continue the effort towards building a successful multimodal assistant agent.
Specifically, \dn{} is designed to include a closer-to-real-world context for a fashion or furniture shopping scenario, moving away from the sanitized contexts present in the original SIMMC datasets.
To this end, we propose a VR scene generator that allows for controlling and capturing diverse multimodal contexts with ground-truth scene graph information, while serving as a close proxy for real-world scenarios.
We then collect $11K$ assistant$\leftrightarrow$user task-oriented dialogs ($117K$ utterances) grounded on diverse photo-realistic VR renders of commercial stores ($1.5K$ different scenes).

The incorporation of the complex and cluttered multimodal contexts introduces several interesting challenges, such as understanding visual \textit{and} dialog coreferences (`\textit{the one directly behind it}', `\textit{the one I mentioned}'), tracking dialog states along with multimodal objects, etc.
In addition, the use of photo-realistic scenes surfaces practical limitations of CV models that need to be addressed, such as the detection of partially observed or obstructed referent objects, the visual texture recognition, \etc.

To this end, we propose four main benchmark tasks that are essential in building a multimodal task assistant: Multimodal Disambiguation, Multimodal Coreference Resolution (MM-Coref), Multimodal Dialog State Tracking (MM-DST), and Response Generation.
We then provide a baseline model trained for these tasks, and highlight the key challenges and future research directions.
}



\section{Related Work}
\label{sec:related_work}

\noindent \textbf{Problem Setup}:
The \dn{} dataset addresses the conversational scenarios where the virtual assistant shares a co-observed scene with a user in addition to the traditional communication that takes place in the form of natural language.
Specifically, we choose the shopping experience as the domain for this study, as it often induces rich multimodal interactions around browsing visually grounded items.
We assume that the assistant agent has ground-truth meta information of every object in the scene, while users only observe those objects through the visual modality to describe and compose a request.
In addition, we allow users to physically navigate within each scene, which we simulate as multiple viewpoints updated at different time steps throughout each dialog.
Thus, models for \dn{} would need to understand the user utterance using both the dialog history and the state of the environment as multimodal context.

\noindent \textbf{Multimodal Dialog Datasets}:
Note that our problem setup for co-observing assistant scenarios allows for more natural multimodal coreferences to be used as part of user-assistant conversations.
The existing literature in multimodal dialogs \cite{avsd-dstc8,visdial,clevr-dialog,guesswhat,talk-the-walk} often posits the roles of a primary and secondary observer, \ie \textit{questioner} and \textit{answerer} similar to the Visual Question Answering \cite{vqa} tasks, hence showing a different distribution of language.

\sm{
\noindent \textbf{Task-oriented Dialog Systems}:
Many datasets have been developed in the past to support various assistant scenarios (\eg booking hotels, reserving hotels) \cite{dstc2, sgd-dst, multiwoz, multiwoz2.1}, defining many challenges in handling user requests under the unimodal dialog setting.
}
Our setup extends many of these challenges studied in the previous literature on task-oriented dialog systems (\eg DST, slot carryovers) to the unique multimodal settings.

\sm{
The most recent thread in building a task-oriented dialog system is to fine-tune an end-to-end system on a large pre-trained causal language model, which achieves the state-of-the-art performance in many metrics \cite{simpletod, soloist, bert-dst-cmu, bert-dst-alexa, DSTC9_SIMMC}.
We follow this line of work and provide a baseline which extends it to accommodate for the multimodal input.
}

\section{SIMMC 2.0 Dataset}
\label{sec:dataset}

\begin{figure}[t]
    \centering
    \includegraphics[width=0.99\columnwidth]{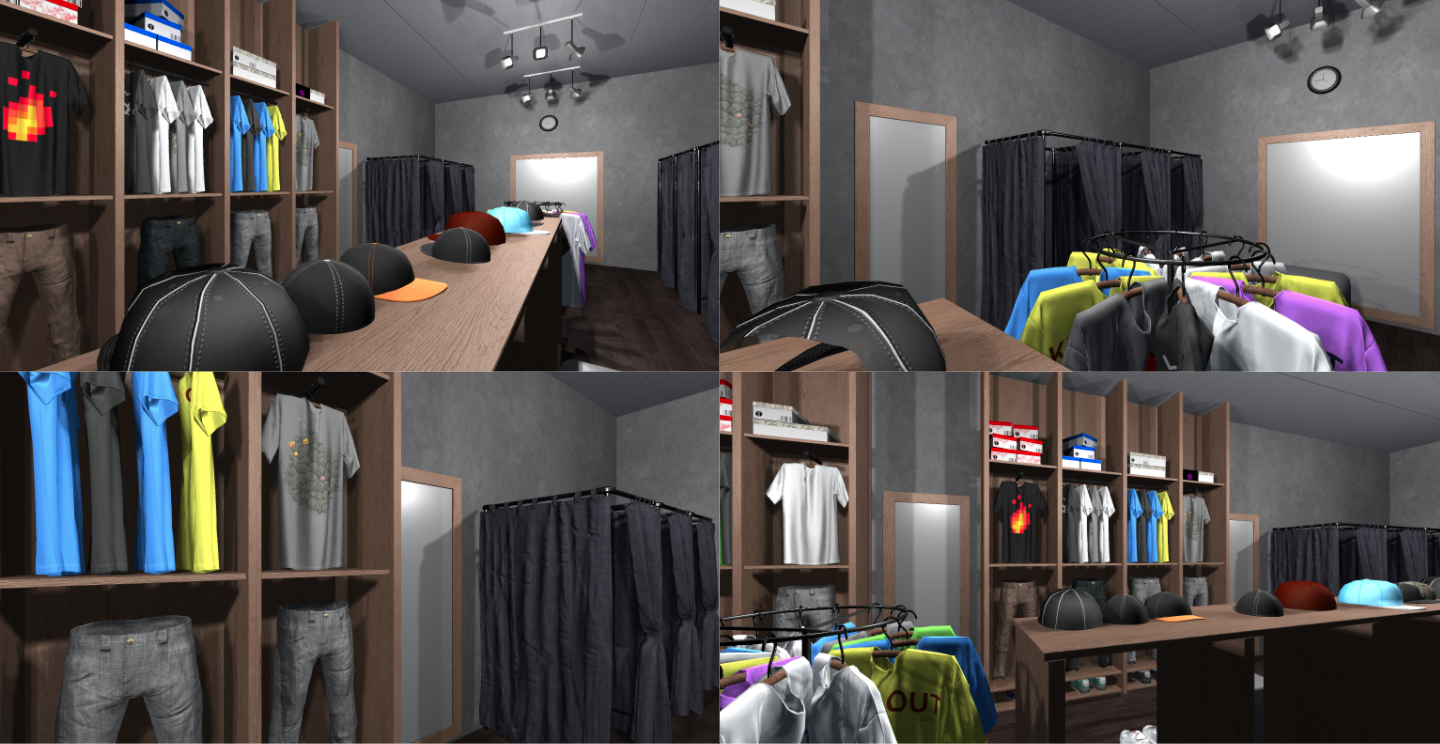}
    \vspace*{\captionReduceTop}
    \caption{
    Example snapshots from random camera viewpoints generated from a rearranged scene.
    Refer to \refsec{sec:scene_simulator} for more details.
    }
    \vspace*{\captionReduceBot}
    \label{fig:scene_generator_examples}
\end{figure}

\begin{figure*}[t]
  \centering \includegraphics[width=1.9\columnwidth]{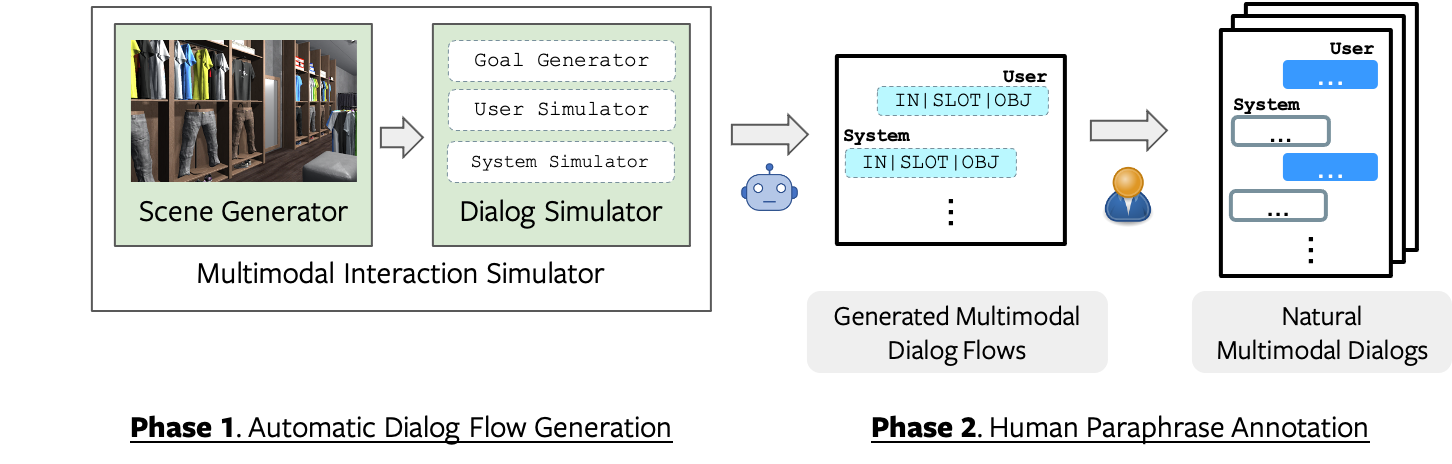}
  \vspace*{\captionReduceTop}
  \caption{Illustration of the two-stage data collection pipeline. \smtodo{expand} Phase 1: Simulated Multimodal Dialog Self-Play (\refsec{subsec:dataset:phase1}) \& Phase 2: Manual Human Paraphrase (\refsec{subsec:dataset:phase2})} 
  \vspace*{\captionReduceBot}
  \label{fig:data_collection}
\end{figure*}

\dn{} assumes the scenario where a user is interacting with a 
conversational assistant to obtain recommendations for a piece of furniture or a clothing item.
The dialogs were collected through a two-phase pipeline (\reffig{fig:data_collection}), 
minimizing the annotation overheads (time and cost).
This approach extends the popular machine$\leftrightarrow$human collaborative dialog collection approaches \cite{sgd-dst, shah2018building} to the multimodal settings.

\subsection{Multimodal Dialog Self-Play}
\label{subsec:dataset:phase1}

\sk{
The first phase entails generating synthetic dialog flows between the user and 
assistant using a multimodal dialog simulator (\refsec{sec:dialog_simulator}).
The simulator conditions the flow generation on various VR scenes snapshots 
(produced by scene simulator) for both fashion and furniture domains.
}
\subsubsection{Scene Simulator}
\label{sec:scene_simulator}
\sk{
In our work, we use photo-realistic, virtual renderings of cluttered shopping
environments for fashion and furniture domains, to replicate real-world settings.
To this end, we develop a scene simulator to generate a diverse set of snapshots
that serve as the multimodal context for the conversations.
}

\noindent
\textbf{Scene Generation.}
\sk{
Re-arranging objects semantically in a 3D environment to create novel arrangements is a long standing
research problem \cite{2012-scenesynth,fisher2015actsynth}.
To avoid the challenges in a completely automatic approach, we design the following pipeline
in a VR environment \cite{unitygameengine}.
We begin by manually constructing photo-realistic shopping scenarios as `seed scenes',
using publicly available digital assets like,
(a) fashion: shirts, dresses, trousers, and shoes,
(b) furniture: sofas, chairs, dining table, and lamps.
We then programmatically re-arrange these assets at random for each of these seed
scenes to create a larger pool of scenes (\reftab{tab:datasets_statistics}),
while keeping the semantics of the scene intact.
For example, a shirt in the seed scene is replaced only by another asset from either the 
same (shirt) or semantic related asset category (\eg, T-shirt, jacket).
This ensures that re-arranged scenes continue to be photo-realistic and avoids object 
collisions and hallucinations, trading off with fixed arrangement
of semantic asset types within each seed scene.
}

\sk{
Finally, we capture multiple views from random camera positions within each scene,
as shown in \reffig{fig:scene_generator_examples}.
The height of the camera is mostly held constant with a small jitter, whereas the 
camera position (when projected onto the floor plane) is randomly chosen and is constrained
to be within $75\%$ of the floor bounds.
These settings give us a good view of the scene objects without the risk of being either:
(a) too close, such that the entire snapshot is taken by partially visible 1--2 objects
resulting in a poor and uninteresting scene view, or,
(b) too far away, where objects are small and hard to differentiate from one another.
Randomly sampled camera viewpoints also encourage the diversity of referring expressions 
(\eg, \textit{`shirt closest to the changing rooms'}, 
\textit{`cap at the farthest end of the table'}),
useful for successful coreferences or disambiguation within the dialog utterances.
}

\noindent
\textbf{Annotation Extraction.}
The synthetic nature of our scenes facilitates an easy extraction of complete scene graph 
information for any given snapshot, without any additional human annotations.
This is particularly beneficial as it enables the generation of rich and interesting dialog 
flows (\refsec{sec:dialog_simulator}),
and allows for a tighter control over the distribution of objects and attributes within 
the conversation, which is nearly impossible with real world multimodal contexts.
The annotations we extract for each scene snapshot consists of all the assets that appear in
the snapshot, their image 2D bounding box, and an index to cross reference additional 
metadata from the catalog (\eg, price, available sizes, color, pattern).
After extracting these annotations, we filter out snapshots with less than 5 objects in 
the field of view, and input the remaining scenes into the dialog simulator.
See \refsec{sec:dataset_analysis} for a detailed analysis of the generated scene snapshots 
and the underlying assets used in our work.

\subsubsection{Multimodal Dialog Simulator}
\label{sec:dialog_simulator}

The multimodal dialog simulator takes generated scenes along with the meta information (objects, locations, and attributes) to create user$\leftrightarrow$assistant dialog flows, following an agenda-based dialog simulator approach \cite{schatzmann2007agenda}.

\sm{
\noindent 
\textbf{Multimodal Dialog Self-play.}
The dialog simulator consists of three main components: the \textit{goal generator}, the \textit{user simulator}, and the \textit{assistant simulator}.
The goal generator randomly selects an agenda for each dialog, which describes a high-level sequence of \textit{goals} within shopping scenarios (\eg, \texttt{BROWSE} $\rightarrow$ \texttt{GET\_INFO} $\rightarrow$ \texttt{REFINE}; 
see \reffig{fig:fashion_act_transitions}).
Given a goal, the user simulator draws a suitable dialog action following a probability distribution, which consists of natural language understanding (NLU) intents (\eg, \texttt{REQUEST:GET}, \texttt{CONFIRM:ADD\_TO\_CART}), 
slots (\eg, color, pattern), and object references.
The assistant simulator then reads the user request, interacts with the multimodal contexts via the simulated API (\eg for looking up the information of an item from the catalog, recommending items from the scene), and responds with natural language generation (NLG) intents, slots and object references.
This dialog self-play repeats until each goal in the agenda is successfully met, or when the dialog reaches the maximum number of turns.
}

\sm{
\noindent
\textbf{Multimodal Dialog Ontology.}
The dialog annotations for \dn{} include the NLU and NLG intent and slot labels, following the conventional approaches for task-oriented dialog systems \cite{multiwoz2.1, sgd-dst, moon2020situated}.
Extending the dialog ontology for the complex multimodal settings, we also annotate the object references with their corresponding IDs as defined by the bounding boxes in each scene, allowing for a seamless annotation of multimodal contexts and language (\eg \textit{`Do you have anything similar to the two middle jackets on the table?'} $\rightarrow$ \texttt{INFORM:GET\_SIMILAR}, \textit{slots: \{type: jacket\}, objects: [0,8]}).
Note also that the same notation is employed to refer to object mentions that are carried over in the dialog context (\eg \textit{`How much is the jacket I mentioned earlier?'} $\rightarrow$ \texttt{INFORM:GET.price}, \textit{objects: [8]}).
This fine-grained and unified ontology allows for the systematic study of the diverse referring expressions (\ie, object mentions) in multimodal dialogs.  
}

\subsection{Manual Paraphrase}
\label{subsec:dataset:phase2}

\smtodo{expand below}

The simulated dialog flows are then paraphrased by human annotators. This helps us draw utterances from the natural language distribution, as expected in a real world application. For this annotation effort, we designed a tool that displays a multimodal scene (generated VR scene screenshot) and a simulated dialog flow, and asked the human annotators to paraphrase the utterance ensuring that critical information such as objects and attributes is retained. An example dialog is shown in Appendix. 

\begin{table}[t]
    \begin{center}
        \scalebox{0.92}{
        \begin{tabular}{lccc}
        \toprule[\heavyrulewidth]
        Total \# dialogs & 11,244 \\
        Total \# utterances & 117,236 \\
        Total \# scene snapshots & 1566 \\
        Avg \# words per user turns 	& 12 \\
        Avg \# words per assistant turns &	13.7 \\
        Avg \# utterances per dialog 	& 10.4 \\
        Avg \# objects mentioned per dialog &	4.7 \\
        Avg \# objects in scene per dialog &	19.7 \\
        \bottomrule[\heavyrulewidth]
        \end{tabular}
        }
    \end{center}
    \vspace*{\captionReduceTop}
    \caption{\textbf{\dn{} Dataset Statistics}}
    \vspace*{\captionReduceBot}
    \label{tab:datasets_statistics}
\end{table}

\textbf{Advantages of the two-stage approach}:
Since paraphrasing synthetic utterances is much faster and less demanding, our approach requires reduced annotation effort. Further, the simulator in Phase 1 provides all annotations for the dialog state and coreferences for free, \ie, without any additional human annotations.

\subsection{SIMMC 2.0 Dataset Analysis}
\label{sec:dataset_analysis}
We analyze our dataset that contains a total of $11.2k$ dialogs (about $117k$ utterances),
split into $7.2k$ and $4k$ dialogs from fashion and furniture domains respectively,
along with the rich annotations from both scene simulator and multimodal
dialog simulator that are extracted automatically without any additional human annotators.
\reftab{tab:datasets_statistics} shows the overall statistics of the dataset.

\begin{figure*}[t]
    \centering
    \begin{subfigure}[b]{0.49\textwidth}
        \includegraphics[width=\textwidth]{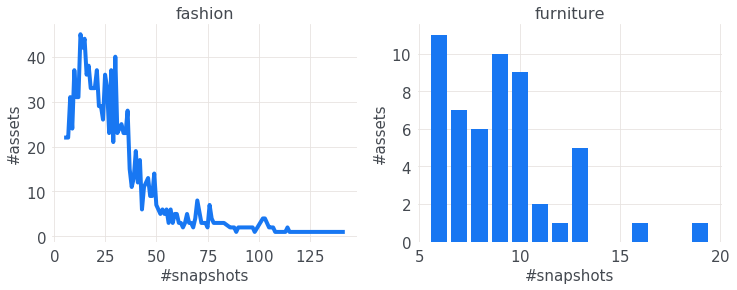}
        \caption{}
        \label{fig:snapshot_asset_distr}
    \end{subfigure}
    ~
    \begin{subfigure}[b]{0.49\textwidth}
        \includegraphics[width=\textwidth]{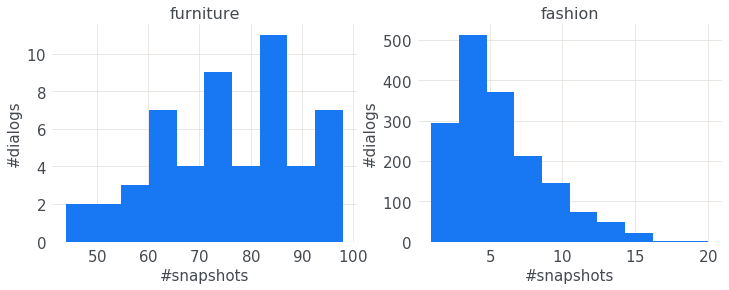}
        \caption{}
        \label{fig:snapshot_dialog_distr}
    \end{subfigure}
    
    \begin{subfigure}[b]{0.23\textwidth}
        \includegraphics[width=\textwidth]{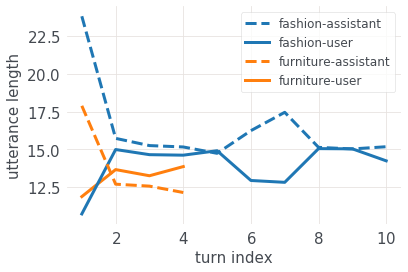}
        \caption{}
        \label{fig:utterance_len_distr}
    \end{subfigure}
    ~
    \begin{subfigure}[b]{0.48\textwidth}
        \includegraphics[width=\textwidth]{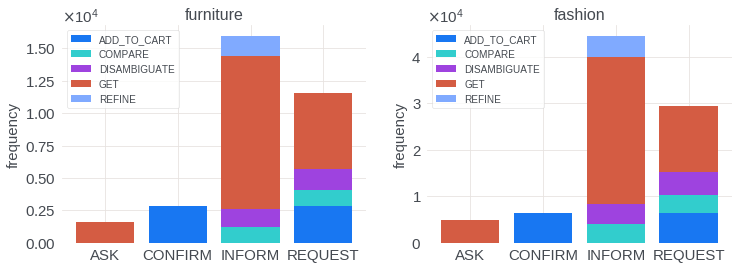}
        \caption{}
        \label{fig:dialog_act_distr}
    \end{subfigure}
    ~
    \begin{subfigure}[b]{0.23\textwidth}
        \includegraphics[width=\textwidth]{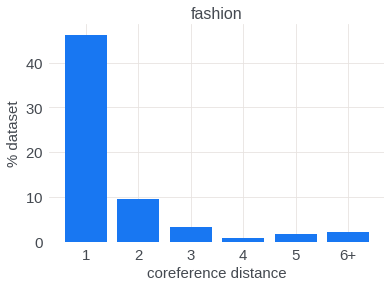}
        \caption{}
        \label{fig:coreference_distance_distr}
    \end{subfigure}
    \vspace*{\captionReduceTop}
    \caption{Distribution of 
    (a) number of assets in each snapshot,
    (b) number of dialogs for each snapshot,
    (c) utterance lengths with dialog turns,
    (d) acts and activities, and
    (e) coreference distance between object mentions.}
    \vspace*{\captionReduceBot}
\end{figure*}

\noindent
\textbf{Analyzing Assets \& Scene Snapshots.}
In our work, we use around $290$ digital assets for 
fashion\footnote{\url{https://www.turbosquid.com/}}, and
$110$ assets for furniture\footnote{\url{https://www.wayfair.com/}}, across several
asset categories shown in \reftab{tab:simmc_assets}.
From these assets, we construct $7$ seed scenes for fashion and $1$ seed scene for furniture.
We then rearrange assets within each seed scenes $20$ times to result in a pool of $140$ and 
$20$ scenes for respective domains.
Finally, we extract $10$ snapshots from random camera viewpoints for each scene in the pool
giving us a total of about $1566$ unique scene snapshots to choose from post filtering.
The number of objects in each snapshot is distributed as shown in
\reffig{fig:snapshot_asset_distr}.
This huge variance presents a good opportunity for the dialog simulator to ground the 
conversational flow in a varied number of objects.

\begin{table*}[th]
    \centering
    \begin{tabular}{l p{5.5in}}
    \toprule
         Fashion &
         hat, tshirt, jacket, hoodie, sweater, shirt, suit, vest, coat, trousers,
         jeans, joggers, skirt, blouse, tank top, dress, shoes \\
         Furniture &
         area rug, bed, chair, couch chair, dining table, coffee table, end table,
         lamp, shelves, sofa \\
    \bottomrule
    \end{tabular}
    \vspace*{\captionReduceTop}
    \caption{Digital Asset Categories used in \dn{} for both fashion and furniture domains.}
    \vspace*{\captionReduceBot}
    \label{tab:simmc_assets}
\end{table*}

\noindent
\textbf{Analyzing Dialog Annotations.}
The dialog simulator generates dialog flows by randomly sampling from the pool of 
$1566$ filtered snapshots, which are later manually paraphrased.
While most of the dialogs ($9.3k$) are grounded in a single snapshot, we include few dialog
flows that spans over two snapshots ($1.9k$) with overlapping set of objects. 
This allows for modeling interesting conversations that require a context carry-over 
across the two viewpoints, thus moving closer to the real-world scenario.
Each snapshot corresponds to about $7.1$ dialogs on an average with the distribution
shown in \reffig{fig:snapshot_dialog_distr}.
Further, each dialog contains around $5.2$ utterance pairs (user$\leftrightarrow$assistant), 
where the utterances are $12.0$ and $13.7$ tokens long respectively
(see \reffig{fig:utterance_len_distr} for distribution over different turns).

\begin{figure*}
    \centering
    \includegraphics[width=0.99\textwidth, trim={0in 0.3in 0in 1in}]{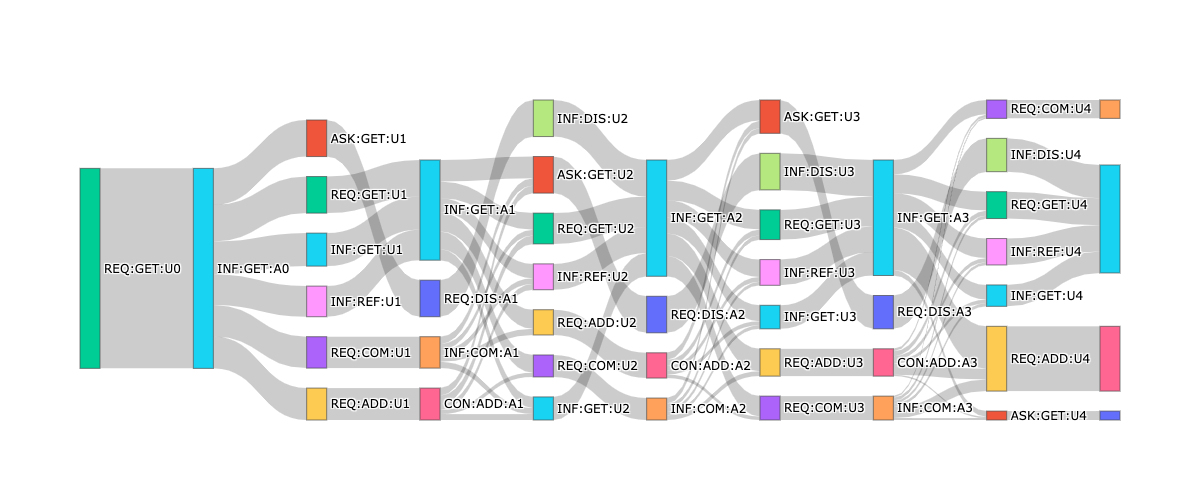}
    \vspace*{\captionReduceTop}
    \vspace*{-20pt}
    \caption{
    \sk{
    Dialog act transitions for the first four rounds of dialogs in the fashion domain.
    Label \texttt{ACT:ACTIVITY:[A$|$U][turn]} denotes the act and activity 
    (shortened for brevity; see \reffig{sec:dataset_analysis} for full names),
    either \textbf{A}ssistant or \textbf{U}ser utterance with turn index.
    The wide branch-offs and inter-connectivity demonstrates the diversity of dialog
    flows generated by our dialog simulator.
    }}
    \vspace*{\captionReduceBot}
    \label{fig:fashion_act_transitions}
\end{figure*}

Following prior work \cite{moon2020situated}, our dialog annotations also comprise 
dialog acts ($4$: \texttt{INFORM}, \texttt{CONFIRM}, \texttt{REQUEST}, \texttt{ASK}) and 
activities ($5$: \texttt{GET}, \texttt{DISAMBIGUATE}, \texttt{REFINE},
\texttt{ADD\_TO\_CART}, \texttt{COMPARE}).
\reffig{fig:dialog_act_distr} shows their frequency breakdown.
We also visualize the dialog act transitions for furniture in
\reffig{fig:fashion_act_transitions} for the first four rounds of the dialog.
The presence of wide branch-offs and inter-connectivity suggests that our simulator is able 
to generate a diverse set of flows, useful to train a robust conversational system.
It is interesting to note that the user utterances (marked with :U) almost always are more
varied than assistant counterparts (marked with :A).
This is probably due to \texttt{INFORM:GET} (fetching information) being a reasonable assistant
response to a large number of user utterance queries.
\reffig{fig:coreference_distance_distr} shows the challenging nature of
coreferences within our dialogs, where we measure the distance to the latest mention of
an object, and requires models to reason across the utterances.
\sk{Finally, \reftab{tab:references} lists the various types of referring expressions in \dn{}, as implicitly controlled by the dialog simulator.}


\begin{table*}[t]
    \begin{center}
        \scalebox{0.75}{
        \setlength\tabcolsep{9pt}
        \begin{tabular}
        {p{0.11\textwidth} p{0.22\textwidth} p{0.84\textwidth}}
        \toprule[\heavyrulewidth]
            \multicolumn{2}{l}{\textbf{Referring Expression Type}}  & 
            \textbf{Examples} \\
        \midrule
            \multirow{7}{*}{Visual} &
            \multirow{2}{*}{Spatial (Absolute)} &
            \textit{`Have you got anything in the same size as \underline{the black top in the middle of the long rack}?'} \\
            {} &
            {} &
            \textit{`Add \underline{the white couch chair sitting on the red rug}.'} \\
            \cmidrule{2-3}
            {} &
            \multirow{2}{*}{Spatial (Relative)} &
            \textit{`Is there an armchair like \underline{the white one closer to us} but from Modern Arts?'} \\
            {} &
            {} &
            \textit{`I like the light grey jacket that is \underline{closer to us on that shelf}, so please put it into my cart.} \\            
            \cmidrule{2-3}
            {} &
            \multirow{3}{*}{Adjectival} &
            (color) \textit{`Anything else like that \underline{blue sweater}?'} \\
            {} &
            {} &
            (shape) \textit{`There's the brown \underline{z-shaped} end table you might enjoy.'} \\                        
            {} &
            {} &
            (multiple) \textit{`I have that \underline{gray wooden table}.'} \\            
        \midrule
        
            \multirow{7}{*}{Textual} &
            \multirow{2}{*}{Noun Phrase (Paraphrase)} &
            U: \textit{`I'm kinda liking that \underline{brown one in the middle row}$^{*}$. Anything similar?'} \newline \hspace{8pt} $\rightarrow$ (\textit{after 2 turns}) U: \textit{`I mean \underline{the one on the left I told you I liked}$^{*}$.'} \\
            \cmidrule{2-3}
            {} &
            \multirow{2}{*}{Noun Phrase (Copy)} &
            A: \textit{`Check out that \underline{brown table in the back right}$^{*}$.'} \newline \hspace{8pt} $\rightarrow$ (\textit{after 2 turns}) U: \textit{`OK. I've decided. Add \underline{the brown table}$^{*}$.'} \\
            \cmidrule{2-3}
            {} &
            \multirow{1}{*}{Slot Carryover} &
            \textit{`Do you have anything \underline{similar} at an affordable price in black and white?'} \\        
            \cmidrule{2-3}
            {} &
            \multirow{2}{*}{Pronoun} &
            A: \textit{What do you think of \underline{the black sweater on the right wall}$^{*}$?} \newline \hspace{8pt} $\rightarrow$ U: \textit{`I like \underline{it}$^{*}$. Add \underline{it}$^{*}$ to my cart.'}  \\                
        \midrule
        
            \multirow{2}{*}{Ambiguous} &
            \multirow{1}{*}{Dialog Context} &
            \textit{`What's the price?'} \\
            \cmidrule{2-3}
            {} &
            \multirow{1}{*}{Visual Context} &
            \textit{`What's the price on the \textit{blue one}?'} \\
        \bottomrule[\heavyrulewidth]
        \end{tabular}
        }
    \end{center}
    \vspace*{\captionReduceTop}
    \caption{
        \underline{Referring expression} types in \dn{} with examples. (* for each row: referring the same object)
    }
    \vspace*{\captionReduceBot}
    \label{tab:references}
\end{table*}


\subsection{Comparison: SIMMC 2.0 vs SIMMC 1.0}
The key differences between \dn{} (ours) and SIMMC 1.0 \cite{moon2020situated}
are (\reffig{fig:teaser}):

\noindent
(a) The multimodal context in SIMMC 1.0 consists of either co-observed images or VR environment,
which are simplistic and sanitized in comparison to real-world scenarios.
For instance, the VR environment in SIMMC-Furniture comprises three slots (left, center, right) 
to populate the catalog items, thus limiting the complexity of referential or disambiguation language.
In contrast, conversations in \dn{} are grounded in photo-realistic scene renderings of
commercial stores that are cluttered and thus closely represent the real-world contexts.

\noindent
(b) The number of objects in the multimodal context for SIMMC 1.0 is capped at 3
compared to $19.7$ on average for \dn.
This allows for richer coreferences, referential expressions, and disambiguation scenarios, 
elevating the role of dialog.

\noindent
(c) Many of the objects in each scene are only partially observed (\eg, blocked by different items or shelves, out of POV frame), which reflects real-world scenes but poses a more challenging problem for the computer vision module.

\section{Task Formulation}
\label{sec:task_formulation}

\begin{table*}[t]
    \begin{center}
        \scalebox{0.78}{
        \setlength\tabcolsep{9pt}
        \begin{tabular}
        {p{0.35\textwidth} p{0.44\textwidth} p{0.33\textwidth}}
        
        \toprule[\heavyrulewidth]
        \textbf{Task Name}  & \textbf{Goal} & \textbf{Evaluation} \\
        \midrule
        
        1. Multimodal Disambiguation 
        & Given user utterances, classify if the assistant should disambiguate in the 
        next turn.
        & Binary classification accuracy \\
        
        \midrule
        
        2. Multimodal Coreference \newline Resolution (MM-Coref) & 
        Given user utterances with object mentions, resolve referent objects to their
        canonical ID(s) as defined by the catalog. 
        & Coref Precision / Recall / F1 \\
        
        \midrule
        
        3. Multimodal Dialog State Tracking \newline (MM-DST)
        & Given user utterances, track user belief states across multiple turns.
        & Intent Accuracy, Slot Precision / Recall / F1 \\
        
        \midrule
        
        4. Response Generation 
        & Given user utterances, ground-truth APIs and ground-truth object IDs, 
        generate Assistant responses or retrieve from a candidate pool.
        & Generation: BLEU;
        
        Retrieval: Accuracy@k, mean reciprocal rank, mean rank \\
        \bottomrule[\heavyrulewidth]
        \end{tabular}
        }
    \end{center}
    \vspace*{\captionReduceTop}
    \caption{
    Proposed tasks and descriptions on our \dn{} dataset. 
    Please see \refsec{sec:task_formulation} for more details.
    }
    \vspace*{\captionReduceBot}
    \label{tab:tracks}
\end{table*}

\sk{
The aim of the SIMMC 2.0 dataset is to emulate futuristic, real-world shopping scenarios
where humans converse with a dialog agent in natural language grounded in a situated
multimodal context.
As a step towards this intelligent conversational agent, we leverage the dialogs and
annotations in our dataset and propose four benchmark tasks (summarized in 
\reftab{tab:tracks}) along with evaluation metrics.
These tasks capture several multimodal conversational reasoning challenges, as elaborated next.
}

\subsection{Multimodal Disambiguation}
\sk{
In a real-world conversation, humans often use shorthands (coreferences) in order to refer
to objects / events that have already been mentioned in the dialog.
While we reserve modeling coreference resolution as a challenging task in 
\refsec{sec:coreference_task},
it is important for the system to recognize ambiguous uses of such coreferences even before
attempting to resolve them.
For example, consider
\textit{`A: The blue trousers are priced at \$45. U: What about \underline{those}?'},
where the phrase \underline{those} could be ambiguous in the following situations:
(a) The user refers to a group of trousers without specifying the exact one they have in mind,
(b) The user incorrectly uses a shorthand for a novel pair of trousers not mentioned in the 
dialog due to conversational brevity.
In either cases, identifying the need for disambiguation and responding with 
\textit{`Which ones are you talking about? The red or the green pair?'} is a desirable
trait for a robust assistant system.
The multimodal disambiguation task tests this ability of the agent.
}

\sk{
More concretely, given the dialog history and the current user utterance, multimodal
disambiguation requires the agent to predict a binary label conditioned on the multimodal
context, to indicate the presence of a referential ambiguity in the user utterance.
This label could also be useful for other downstream tasks like assistant response
generation (\refsec{sec:response_generation_task}) in order to continue the conversation 
in a meaningful way. 
We use accuracy to measure and compare model performances for this task.
}

\subsection{Multimodal Coreference Resolution} 
\label{sec:coreference_task}

For this task, we aim to resolve referential mentions in user utterances to their canonical object IDs as defined for each scene.
These mentions can be resolved through (1) the dialog context (\eg A: \textit{`\underline{This shirt} comes in XL and is \$29.'} $\rightarrow$ U: \textit{`Please add \underline{it} to cart.'}, or (2) the multimodal context (\eg U: \textit{`How much is that \underline{red shirt}?}'), or (3) both (\eg U: \textit{`How much is \underline{the one} next to \underline{the one you mentioned}?'}).

\sm{
The input for this task includes the ground-truth bounding boxes defining each object ID, 
to avoid the performance bottleneck by the object detection algorithms.
The main evaluation metric includes F1, precision and recall performance.
Note that we exclude from evaluation the object mentions that are immediately followed by a disambiguation request (\eg, \textit{`How much is \underline{the one over there}?' $\leftrightarrow$ \textit{`Which one do you mean?'}}, as they provide insufficient 
descriptions for resolving those coreferences.}

\subsection{Multimodal Dialog State Tracking} 
Following \citet{moon2020situated}, we extend the traditional notion of the unimodal dialog state
tracking (DST) and propose multimodal dialog state tracking (MM-DST) as a main sub-task where slots are grounded on the coexisting multimodal context, which requires handling of multimodal objects (as opposed to textual tokens) as part of dialog states.

\sm{
The performance is measured by the joint F1, recall and precision performance for the cumulative intent, slot and object reference predictions. 
The underlying reasoning behind this task is that the MM-DST labels will be able to provide sufficient information for a multimodal dialog system to carry out dialog policies and actions, given the detected and resolved items in each multimodal scene.
Therefore, the MM-DST task measures the model's holistic understanding of user requests throughout each dialog, including the disambiguation needs as well as the coreferences.
}

\subsection{Assistant Response Generation}
\label{sec:response_generation_task}

\sk{
The goal of this task is to generate assistant responses or retrieve from a candidate pool, given user utterances, ground-truth belief state, and object IDs.
While we assume the assistant agent has the ground-truth meta information on each object, each
response needs to naturally describe the referent objects \textit{as observed and understood} by
the user through the co-observed scene or the dialog context 
(\eg \texttt{INFORM:RECOMMEND (OBJ\_ID: 3)} $\rightarrow$ A: ``\textit{I recommend the blue shirt directly behind the brown jacket.}".

Similar to \cite{moon2020situated}, we propose two ways to evaluate the performance of
systems for response generation:
(a) As a \textbf{generation} task, where the agent is seen as conditional language
model.
Performance is measured using BLEU-4 score \cite{papineni-etal-2002-bleu} between the
generated response and the ground truth response provided with the dataset.
(b) As a \textbf{retrieval} task, where the agent has to pick the ground truth response 
from a list of candidate responses (generated randomly; unique to each utterance).
We use traditional information retrieval metrics like recall@k ($k=\{1, 5, 10\}$),
mean rank, and mean reciprocal rank for comparing model performances.
}

\section{Modeling \& Empirical Analysis}
\label{sec:model_analysis}
In this section, we perform preliminary empirical analysis and train baselines.
We leave more detailed modeling work for the future.

\noindent \textbf{Dataset split}.
We randomly split the dataset into 4 sets: train ($65\%$), dev ($5\%$), 
dev-test ($15\%$), and test-std ($15\%$), which we leave as a held-out hidden set for performing a fair comparison of models.

\begin{table}[t]
    \setlength{\tabcolsep}{4pt}
    \begin{center}
        \scalebox{0.72}{%
            \begin{tabular}{
                ccccc
            }
            \toprule[\heavyrulewidth]
            \multicolumn{1}{c}{\textbf{1. Disamb.}}
            & \multicolumn{1}{c}{\textbf{2. MM-Coref}}
            & \multicolumn{2}{c}{\textbf{3. DST}}
            & \multicolumn{1}{c}{\textbf{4. Gen.}}\\            
            \cmidrule(r){1-1}
            \cmidrule(r){2-2}
            \cmidrule(r){3-4}
            \cmidrule(r){5-5}
                Acc$\uparrow$
                & Coref F1$\uparrow$
                & Slot F1$\uparrow$
                & Intent F1$\uparrow$
                & BLEU$\uparrow$ \\            
            \midrule
                \reportval{73.9}{1.2}
                & \reportval{36.64}{0.58}
                & \reportval{81.72}{0.51}
                & \reportval{94.53}{0.36}
                & \reportval{0.192}{0.002}\\
            \midrule
                -
                & -
                & \reportval{74.75}{0.42}
                & \reportval{93.40}{0.26}
                & \reportval{0.217}{0.002}\\            
            \bottomrule[\heavyrulewidth]
            \end{tabular}
        }
    \end{center}
    \vspace*{\captionReduceTop}
        \caption{
        Baseline performances: \citet{moon2020situated} (top), \citet{le2019multimodal} (bottom).
        \textbf{(1) Multimodal Disambiguation (Disamb.)}, via classification
        \underline{accurac}y, 
        \textbf{(2) Multimodal Coreference Resolution (MM-Coref)}, via \underline{coref} prediction \underline{F1},
        \textbf{(3) Dialog State Tracking (DST)}, via \underline{slot} and \underline{intent} \underline{F1},
        \textbf{(4) Response Generation} via \underline{BLEU}.
        $\uparrow$: higher is better.
        }
    \vspace*{\captionReduceBot}
    \label{tab:results}    
\end{table}

\sm{
\noindent \textbf{Notations.}
We denote a \simmc dialog with $N_r$ rounds: $\mathcal{D} = \{(U_i, A_i, M_i, B_i)\}_{i=1}^{N_r}$,
where $U_i$ and $A_i$ are the user and assistant utterances,
$M_i$ is the domain-specific multimodal context,
and $B_i$ is a multimodal belief state represented as a semantic parse of user-side dialog (\ie intent, slot, object references, disambiguation labels), respectively.
At each round $t$, given the current user utterance $U_t$, the dialog history 
$H_t={(U_i, A_i)}_{i=1}^{t-1}$, and the multimodal context $M_t$,
the task is to predict the user belief state $B_t$, as well as the natural language assistant response $A_t$.
}

\begin{figure}[t]
  \centering \includegraphics[width=1.0\columnwidth]{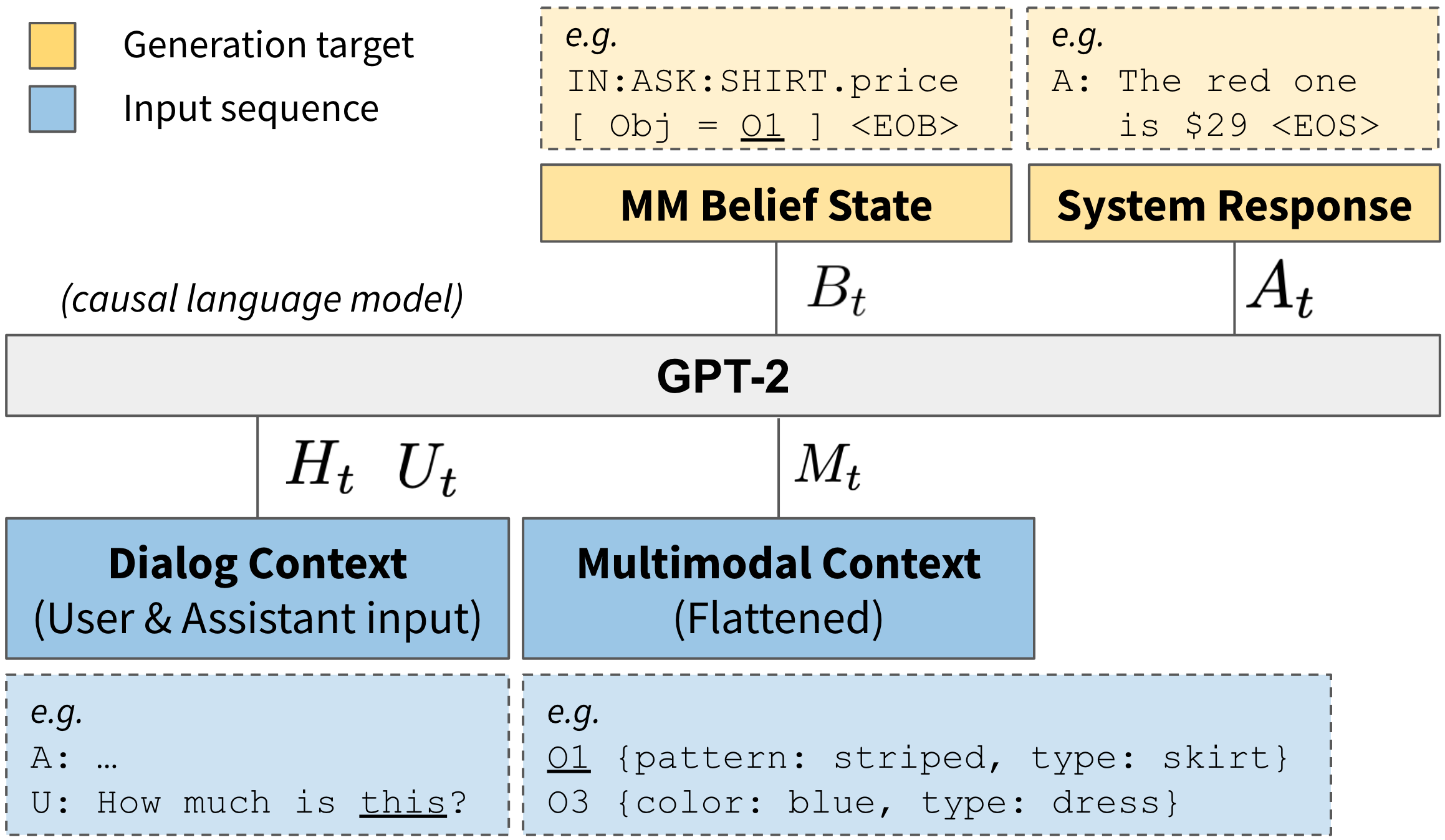}
  \caption{Illustration of the GPT-2 based baseline, which takes as input the dialog context and the flattened multimodal context, and outputs the belief states as well as the system response. } 
  \vspace*{\captionReduceTop}
  \label{fig:model_architecture}
  \vspace*{\captionReduceBot}
\end{figure}

\noindent \textbf{Baselines.}
We benchmark the dataset by adopting:
(a) \textbf{MM-DST model} by \citet{moon2020situated}, where we train a multi-task GPT-2~\cite{radford2019language} based Transformer model using the joint supervision signals for the Disambiguation, MM-Coref, DST, and Response Generation tasks.
Specifically, the model takes as input the dialog context and the flattened
multimodal contexts (as structurally formatted strings) to predict the belief states and
the responses, following the popular causal language model approach \cite{soloist, simpletod}.
We use the $12$-layer GPT-2 ($117$M parameters) as the pre-trained language model and fine-tune for ten epochs. 
Note that this baseline uses the ground-truth multimodal contexts provided from the scene generator, 
instead of consuming raw images as input, and thus serves as a soft oracle on the proposed dataset.
(b) \textbf{Multimodal Transformer Network (MTN)} \cite{le2019multimodal} for the DST and 
Response Generation tasks.
In particular, MTN uses image features extracted from scene snapshots and attends to relevant
parts as guided by the dialog.
We use the same training setting and hyperparameters as \citet{le2019multimodal}.

\noindent \textbf{Analysis.}
The results are summarized in \reftab{tab:results}.
Note that the F1 performance on the multimodal object coreference resolution task on \dn{} is only at 36.6\%, whereas the best model on SIMMC 1.0 \cite{moon2020situated} achieved 85.9\% on the similar task.
This demonstrates that \dn{} presents more complex and cluttered scenes, thus requires more rigorous visual grounding of multimodal contexts (19.7 objects per dialog on average).


\noindent \textbf{Conclusions.}
We present a novel dataset for the Situated and Interactive Multimodal Conversations,
\dn{}, with $11K$ user$\leftrightarrow$assistant dialogs ($117K$ utterances) on shopping domain (fashion and furniture), grounded in situated and photo-realistic VR scenes.
We then present a novel multimodal dialog simulator, which generates simulated dialogs grounded on diverse multimodal contexts that are automatically configured.
Our empirical analysis with a baseline model demonstrates many new challenges that our \dn{} dataset brings, highlighting new directions of research in this area.

\bibliography{bibliography}

\begin{thebibliography}{30}
\expandafter\ifx\csname natexlab\endcsname\relax\def\natexlab#1{#1}\fi

\bibitem[{Antol et~al.(2015)Antol, Agrawal, Lu, Mitchell, Batra,
  Lawrence~Zitnick, and Parikh}]{vqa}
Stanislaw Antol, Aishwarya Agrawal, Jiasen Lu, Margaret Mitchell, Dhruv Batra,
  C~Lawrence~Zitnick, and Devi Parikh. 2015.
\newblock {VQA}: Visual question answering.
\newblock In \emph{ICCV}.

\bibitem[{Budzianowski et~al.(2018)Budzianowski, Wen, Tseng, Casanueva, Ultes,
  Ramadan, and Ga{\v{s}}i{\'c}}]{multiwoz}
Pawe{\l} Budzianowski, Tsung-Hsien Wen, Bo-Hsiang Tseng, I{\~n}igo Casanueva,
  Stefan Ultes, Osman Ramadan, and Milica Ga{\v{s}}i{\'c}. 2018.
\newblock {M}ulti{WOZ} - a large-scale multi-domain wizard-of-{O}z dataset for
  task-oriented dialogue modelling.
\newblock In \emph{Proceedings of the Conference on Empirical Methods in
  Natural Language Processing (EMNLP)}.

\bibitem[{Chao and Lane(2019)}]{bert-dst-cmu}
Guan-Lin Chao and Ian Lane. 2019.
\newblock Bert-dst: Scalable end-to-end dialogue state tracking with
  bidirectional encoder representations from transformer.
\newblock In \emph{Annual Conference of the International Speech Communication
  Association (INTERSPEECH)}.

\bibitem[{Crook et~al.(2021)Crook, Kottur, Moon, Beirami, Cho, Subba, and
  Geramifard}]{DSTC9_SIMMC}
Paul~A. Crook, Satwik Kottur, Seungwhan Moon, Ahmad Beirami, Eunjoon Cho, Rajen
  Subba, and Alborz Geramifard. 2021.
\newblock Situated interactive multimodal conversations (simmc) track at dstc9.
\newblock \emph{AAAI DSTC9 Workshop}.

\bibitem[{Das et~al.(2017)Das, Kottur, Gupta, Singh, Yadav, Moura, Parikh, and
  Batra}]{visdial}
Abhishek Das, Satwik Kottur, Khushi Gupta, Avi Singh, Deshraj Yadav,
  Jos{\'e}~MF Moura, Devi Parikh, and Dhruv Batra. 2017.
\newblock Visual dialog.
\newblock In \emph{CVPR}.

\bibitem[{Eric et~al.(2019)Eric, Goel, Paul, Kumar, Sethi, Ku, Goyal, Agarwal,
  Gao, and Hakkani-Tur}]{multiwoz2.1}
Mihail Eric, Rahul Goel, Shachi Paul, Adarsh Kumar, Abhishek Sethi, Peter Ku,
  Anuj~Kumar Goyal, Sanchit Agarwal, Shuyag Gao, and Dilek Hakkani-Tur. 2019.
\newblock Multiwoz 2.1: Multi-domain dialogue state corrections and state
  tracking baselines.
\newblock \emph{arXiv preprint arXiv:1907.01669}.

\bibitem[{Fisher et~al.(2012)Fisher, Ritchie, Savva, Funkhouser, and
  Hanrahan}]{2012-scenesynth}
Matthew Fisher, Daniel Ritchie, Manolis Savva, Thomas Funkhouser, and Pat
  Hanrahan. 2012.
\newblock Example-based synthesis of 3d object arrangements.
\newblock In \emph{ACM SIGGRAPH Asia 2012 papers}, SIGGRAPH Asia '12.

\bibitem[{Fisher et~al.(2015)Fisher, Savva, Li, Hanrahan, and
  Nie{\ss}ner}]{fisher2015actsynth}
Matthew Fisher, Manolis Savva, Yangyan Li, Pat Hanrahan, and Matthias
  Nie{\ss}ner. 2015.
\newblock Activity-centric scene synthesis for functional 3d scene modeling.
\newblock \emph{ACM Transactions on Graphics (TOG)}, 34(6).

\bibitem[{Gao et~al.(2019)Gao, Abhishek Seth~and, Chun, and
  Hakkani-Ture}]{bert-dst-alexa}
Shuyang Gao, Sanchit~Agarwal Abhishek Seth~and, Tagyoung Chun, and Dilek
  Hakkani-Ture. 2019.
\newblock Dialog state tracking: A neural reading comprehension approach.
\newblock In \emph{Special Interest Group on Discourse and Dialogue (SIGDIAL)}.

\bibitem[{Gunasekara et~al.(2020)Gunasekara, Kim, D'Haro, Rastogi, Chen, Eric,
  Hedayatnia, Gopalakrishnan, Liu, Huang et~al.}]{dstc9}
Chulaka Gunasekara, Seokhwan Kim, Luis~Fernando D'Haro, Abhinav Rastogi,
  Yun-Nung Chen, Mihail Eric, Behnam Hedayatnia, Karthik Gopalakrishnan, Yang
  Liu, Chao-Wei Huang, et~al. 2020.
\newblock Overview of the ninth dialog system technology challenge: Dstc9.
\newblock \emph{arXiv preprint arXiv:2011.06486}.

\bibitem[{Henderson et~al.(2014)Henderson, Thomson, and Williams}]{dstc2}
Matthew Henderson, Blaise Thomson, and Jason~D Williams. 2014.
\newblock The second dialog state tracking challenge.
\newblock In \emph{Proceedings of the 15th annual meeting of the special
  interest group on discourse and dialogue (SIGDIAL)}, pages 263--272.

\bibitem[{Hori et~al.(2018)Hori, Cherian, Marks, and Metze}]{avsd-dstc8}
Chiori Hori, Anoop Cherian, Tim~K. Marks, and Florian Metze. 2018.
\newblock Audio visual scene-aware dialog track in dstc8.
\newblock \emph{DSTC Track Proposal}.

\bibitem[{Hosseini-Asl et~al.(2020)Hosseini-Asl, McCann, Wu, Yavuz, and
  Socher}]{simpletod}
Ehsan Hosseini-Asl, Bryan McCann, Chien-Sheng Wu, Semih Yavuz, and Richard
  Socher. 2020.
\newblock A simple language model for task-oriented dialogue.
\newblock \emph{arXiv preprint arXiv:2005.00796}.

\bibitem[{Huang et~al.(2021)Huang, Tan, Ng, Shi, Yeo, Jiang, and
  Kim}]{simmc1-team4}
Xin Huang, Chor~Seng Tan, Yan~Bin Ng, Wei Shi, Kheng~Hui Yeo, Ridong Jiang, and
  Jung~Jae Kim. 2021.
\newblock Joint generation and bi-encoder for situated interactive multimodal
  conversations.
\newblock \emph{AAAI 2021 DSTC9 Workshop}.

\bibitem[{Jeong et~al.(2021)Jeong, Lee, Ko, and Seo}]{simmc1-team3}
Younghoon Jeong, Se~Jin Lee, Youngjoong Ko, and Jungyun Seo. 2021.
\newblock Tom : End-to-end task-oriented multimodal dialog system with gpt-2.
\newblock \emph{AAAI 2021 DSTC9 Workshop}.

\bibitem[{Kim et~al.(2021)Kim, Lee, Jeong, Youngjoong, Koo, and
  Seo}]{simmc1-team2}
Byoungjae Kim, Inkwon Lee, Yeonseok Jeong, Ko~Youngjoong, Myoung-Wan Koo, and
  Jungyun Seo. 2021.
\newblock Improving multimodal api prediction via adding dialog state and
  various multimodal gates.
\newblock \emph{AAAI 2021 DSTC9 Workshop}.

\bibitem[{Kottur et~al.(2019)Kottur, Moura, Parikh, Batra, and
  Rohrbach}]{clevr-dialog}
Satwik Kottur, Jos{\'e}~MF Moura, Devi Parikh, Dhruv Batra, and Marcus
  Rohrbach. 2019.
\newblock Clevr-dialog: A diagnostic dataset for multi-round reasoning in
  visual dialog.
\newblock \emph{arXiv preprint arXiv:1903.03166}.

\bibitem[{Kung et~al.(2021)Kung, Yang, Chang, Hsu, Liou, and
  Chen}]{simmc1-team1}
Po-Nien Kung, Tse-Hsuan Yang, Chung-Cheng Chang, Hsin-Kai Hsu, Yu-Jia Liou, and
  Yun-Nung Chen. 2021.
\newblock Multi-task learning for situated multi-domain end-to-end dialogue
  systems.
\newblock \emph{AAAI 2021 DSTC9 Workshop}.

\bibitem[{Le et~al.(2019)Le, Sahoo, Chen, and Hoi}]{le2019multimodal}
Hung Le, Doyen Sahoo, Nancy Chen, and Steven Hoi. 2019.
\newblock Multimodal transformer networks for end-to-end video-grounded
  dialogue systems.
\newblock In \emph{Proceedings of the 57th Annual Meeting of the Association
  for Computational Linguistics}, pages 5612--5623.

\bibitem[{Moon et~al.(2020)Moon, Kottur, Crook, De, Poddar, Levin, Whitney,
  Difranco, Beirami, Cho, Subba, and Geramifard}]{moon2020situated}
Seungwhan Moon, Satwik Kottur, Paul~A Crook, Ankita De, Shivani Poddar,
  Theodore Levin, David Whitney, Daniel Difranco, Ahmad Beirami, Eunjoon Cho,
  Rajen Subba, and Alborz Geramifard. 2020.
\newblock Situated and interactive multimodal conversations.
\newblock \emph{arXiv preprint arXiv:2006.01460}.

\bibitem[{Papineni et~al.(2002)Papineni, Roukos, Ward, and
  Zhu}]{papineni-etal-2002-bleu}
Kishore Papineni, Salim Roukos, Todd Ward, and Wei-Jing Zhu. 2002.
\newblock \href {https://doi.org/10.3115/1073083.1073135} {{B}leu: a method for
  automatic evaluation of machine translation}.
\newblock In \emph{Proceedings of the 40th Annual Meeting of the Association
  for Computational Linguistics}, pages 311--318, Philadelphia, Pennsylvania,
  USA. Association for Computational Linguistics.

\bibitem[{Peng et~al.(2020)Peng, Li, Li, Shayandeh, Liden, and Gao}]{soloist}
Baolin Peng, Chunyuan Li, Jinchao Li, Shahin Shayandeh, Lars Liden, and
  Jianfeng Gao. 2020.
\newblock Soloist: Few-shot task-oriented dialog with a single pre-trained
  auto-regressive model.
\newblock \emph{arXiv preprint arXiv:2005.05298}.

\bibitem[{Radford et~al.(2019)Radford, Wu, Child, Luan, Amodei, and
  Sutskever}]{radford2019language}
Alec Radford, Jeffrey Wu, Rewon Child, David Luan, Dario Amodei, and Ilya
  Sutskever. 2019.
\newblock Language models are unsupervised multitask learners.
\newblock \emph{OpenAI blog}, 1(8):9.

\bibitem[{Rastogi et~al.(2019)Rastogi, Zang, Sunkara, Gupta, and
  Khaitan}]{sgd-dst}
Abhinav Rastogi, Xiaoxue Zang, Srinivas Sunkara, Raghav Gupta, and Pranav
  Khaitan. 2019.
\newblock Towards scalable multi-domain conversational agents: The
  schema-guided dialogue dataset.
\newblock In \emph{Association for the Advancement of Artificial Intelligence
  (AAAI)}.

\bibitem[{Schatzmann et~al.(2007)Schatzmann, Thomson, Weilhammer, Ye, and
  Young}]{schatzmann2007agenda}
Jost Schatzmann, Blaise Thomson, Karl Weilhammer, Hui Ye, and Steve Young.
  2007.
\newblock Agenda-based user simulation for bootstrapping a pomdp dialogue
  system.
\newblock In \emph{Human Language Technologies 2007: The Conference of the
  North American Chapter of the Association for Computational Linguistics;
  Companion Volume, Short Papers}, pages 149--152.

\bibitem[{Senese et~al.(2021)Senese, Rizzo, Benincasa, and
  Caputo}]{simmc1-team5}
Matteo~Antonio Senese, Giuseppe Rizzo, Alberto Benincasa, and Barbara Caputo.
  2021.
\newblock A response retrieval approach for dialogue using a multi-attentive
  transformer.
\newblock \emph{AAAI 2021 DSTC9 Workshop}.

\bibitem[{Shah et~al.(2018)Shah, Hakkani-T{\"u}r, T{\"u}r, Rastogi, Bapna,
  Nayak, and Heck}]{shah2018building}
Pararth Shah, Dilek Hakkani-T{\"u}r, Gokhan T{\"u}r, Abhinav Rastogi, Ankur
  Bapna, Neha Nayak, and Larry Heck. 2018.
\newblock Building a conversational agent overnight with dialogue self-play.
\newblock \emph{arXiv preprint arXiv:1801.04871}.

\bibitem[{{Unity Technologies}(2019)}]{unitygameengine}
{Unity Technologies}. 2019.
\newblock Unity.
\newblock https://unity.com/.

\bibitem[{de~Vries et~al.(2018)de~Vries, Shuster, Batra, Parikh, Weston, and
  Kiela}]{talk-the-walk}
Harm de~Vries, Kurt Shuster, Dhruv Batra, Devi Parikh, Jason Weston, and Douwe
  Kiela. 2018.
\newblock Talk the walk: Navigating new york city through grounded dialogue.
\newblock \emph{arXiv preprint arXiv:1807.03367}.

\bibitem[{de~Vries et~al.(2017)de~Vries, Strub, Chandar, Pietquin, Larochelle,
  and Courville}]{guesswhat}
Harm de~Vries, Florian Strub, Sarath Chandar, Olivier Pietquin, Hugo
  Larochelle, and Aaron Courville. 2017.
\newblock Guesswhat?! visual object discovery through multi-modal dialogue.
\newblock In \emph{CVPR}.

\end{thebibliography}
 \bibliographystyle{acl_natbib}

\begin{figure*}[b]
  \centering \includegraphics[width=1.55\columnwidth]{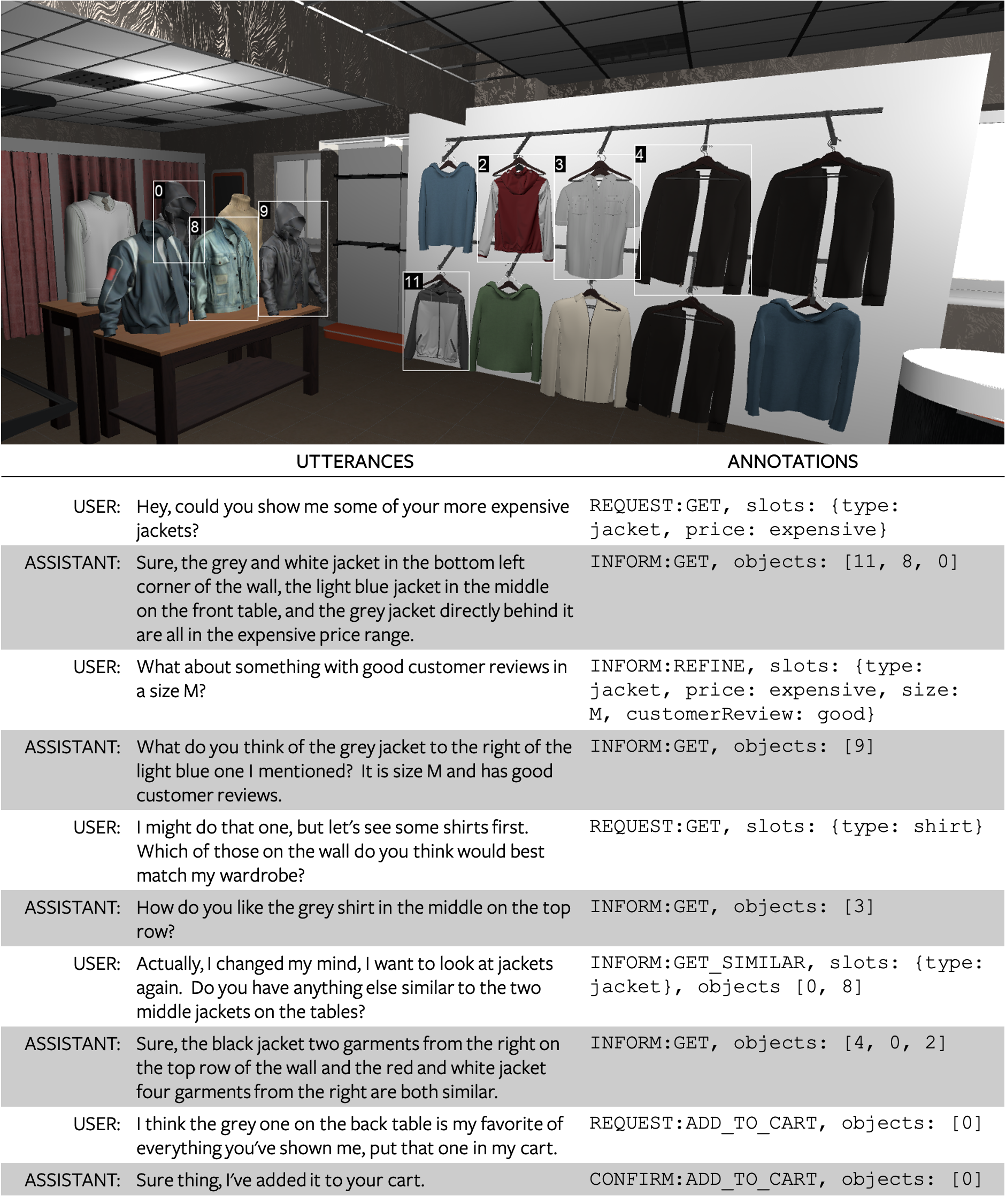}
    \caption{\textbf{Dataset Example}. Dialog labels include intent, slots, and multimodal coreferences. The bounding boxes for the objects mentioned in the dialog (and their corresponding object IDs) are shown for readability.}
  \label{fig:example}
\end{figure*}


\end{document}